\documentclass{article}

\usepackage[preprint]{neurips_2026}

\workshoptitle{Transitioning from Pre-Training to Post-Training}

\usepackage[utf8]{inputenc} 
\usepackage[T1]{fontenc}    
\usepackage{hyperref}       
\usepackage{url}            
\usepackage{booktabs}       
\usepackage{amsfonts}       
\usepackage{nicefrac}       
\usepackage{microtype}      
\usepackage{xcolor}         
\usepackage{amsmath}
\usepackage{amssymb}
\usepackage{booktabs}
\usepackage{tabularx}
\usepackage{pgfplots}
\usepackage{mathtools}
\usepackage{adjustbox}
\usepackage{tikz}
\usepackage{subcaption}
\usetikzlibrary{calc}
\usepgfplotslibrary{groupplots}

\title{Diagnosing and Stabilizing RLVR Training through Entropy Dynamics}

\author{
\begin{tabular}{cccc}
Prasanth YSS & Zhichen Ren & Guangwei Yu & Satya Krishna Gorti \\[0.5em]
\multicolumn{4}{c}{\texttt{\{prasanth, zhichen, guang, satya\}@layer6.ai}} \\[1.5em]
\multicolumn{4}{c}{Layer 6 AI} 
\end{tabular}
}

\begin{document}

\maketitle

\begin{abstract}
    Reinforcement Learning with Verifiable Rewards (RLVR) is a central component of modern LLM post-training, yet the training is prone to instability and collapse. Prior work has largely studied this instability through the lens of off-policiness, but the resulting mitigation strategies do not consistently stabilize training across settings. Rather than attributing instability to a particular training–inference mismatch, we take a complementary approach that characterizes and directly controls a common signature of collapse. Across models and datasets, we find that RLVR collapse is consistently associated with a sharp increase in entropy. We use a first-order estimate of entropy change from the policy gradient and show that, it follows the same qualitative trend as the observed entropy prior to collapse. Directly controlling updates according to this quantity produces consistent improvements in training stability across models and datasets. We further find that the effectiveness of existing stabilization strategies is closely associated with their ability to control this entropy-gradient signal, providing a common explanation for when these interventions succeed or fail. Together, these results provide an entropy-based perspective on RLVR instability and suggest that controlling entropy growth is a key ingredient for stable training.
  
\end{abstract}

\section{Introduction}

Reinforcement learning with verifiable rewards (RLVR) ~\cite{guo2025deepseek} has emerged as a scalable post-training paradigm for language models. Widely used methods such as Group Relative Policy Optimization (GRPO) ~\citet{shao2024deepseekmath} sample a group of responses for each prompt and optimize their group-normalized advantages:

\begin{equation}\label{eq:grpo}
J_{\mathrm{GRPO}}(\theta) = \mathbb{E}_{\substack{q\sim\mathcal{D}\\[-2pt] x\sim\pi_{\theta_{\mathrm{old}}}(\cdot\mid q)}}
\left[\frac{1}{G}\sum_{i=1}^{G}\frac{1}{|x_i|}
\sum_{t=1}^{|x_i|}
\min\left(\rho_{i,t}(\theta)\hat{A}_i,\, \operatorname{clip}\left(
\rho_{i,t}(\theta),1-\epsilon,1+\epsilon
\right)\hat{A}_i
\right)
\right]
\end{equation}

Here, $\rho_{i,t}(\theta)=\pi_\theta(x_{i,t}\mid q,x_{i,<t})/\pi_{\theta_{\mathrm{old}}}(x_{i,t}\mid q,x_{i,<t})$ is the importance ratio. But RLVR training under GRPO is prone to instability. One line of work studies this instability through the mismatch between the surrogate objective in Eq.~\ref{eq:grpo} and the expected reward objective $J(\theta)=\mathbb{E}_{q\sim\mathcal{D},x\sim\pi_\theta(\cdot\mid q)}[R(q,x)]$. \cite{zheng2025groupsequencepolicyoptimization} replaces token-level importance sampling with a sequence-level ratio, \cite{huang2026stable} adapts the update magnitude for unreliable off-policy samples. Other work addresses the consequences of policy mismatch through controlling \textit{gradient-alignment} \cite{xu2026gac}, replaying MoE routing decisions \cite{ma2025stabilizing}, or improving numerical consistency between training and inference \cite{qi2025defeating}. While these approaches improve stability in some settings, controlling policy mismatch does not consistently prevent collapse \cite{park2025clip, yss2026gradient}. Moreover, removing policy lag or train--inference mismatch can be undesirable for efficiency. This motivates studying instability without explicitly modeling the source of policy drift.

A second line of work studies the characteristics of individual token updates. \cite{wang2026beyond} train only on the highest-entropy $20\%$ of tokens, while \cite{yss2026gradient} analyze token-level gradient updates and improves stability by removing negative-advantage updates. These methods identify subsets of updates that improve training, but do not directly explain why the effectiveness of same strategy may vary across settings.

We take an alternative approach and study a common effect associated with RLVR instability. In Section \ref{sec:entropy-dynamics}, we empirically show that collapse is typically associated with an increase in the expected entropy of generated rollouts under the trainer policy. We further show that the accumulated entropy-change estimate in Eq.~\ref{eq:delta_entropy} trends with the observed entropy prior to collapse across datasets, models, and clipping configurations (Fig.~\ref{fig:main-training-results}). In Section~\ref{sec:entropy-thresholding}, we then directly intervene on this quantity. Masking updates with positive predicted entropy change consistently improves stability, while masking entropy-decreasing updates in collapse-prone settings accelerates collapse.

Finally, in Section~\ref{sec:existing-methods}, we revisit existing stabilization methods through the same entropy-gradient lens. Selecting the highest-entropy $20\%$ of tokens~\cite{wang2026beyond} consistently reduces entropy growth and improves stability, although it remains less stable than directly masking updates according to $\widehat{\Delta H}$. We also revisit the peak--valley intervention of WAPO~\cite{yss2026gradient}: peak-only training accumulates positive predicted entropy change and collapses, whereas valley-only training accumulates negative predicted entropy change and remains stable. Together, these results place seemingly distinct stabilization strategies under a common entropy-control framework, where their effectiveness can be understood through their ability to regulate the entropy-gradient signal.


\section{Entropy Dynamics of Token-Level Policy Updates}
\label{sec:entropy-dynamics}


\pgfplotsset{compat=1.18}

\definecolor{qwenott}{HTML}{56B4E9}
\definecolor{qwennumina}{HTML}{D55E00}
\definecolor{gemmamath}{HTML}{009E73}
\definecolor{gemmanumina}{HTML}{CC79A7}

\pgfplotsset{
  plotbase/.style={
    width=5.0cm,
    height=3.0cm,
    grid=major,
    grid style={draw=black!12},
    axis line style={draw=black!65},
    tick label style={font=\small},
    label style={font=\small},
    scaled ticks=false,
    unbounded coords=discard,
    legend style={
      font=\scriptsize,
      draw=none,
      fill=none,
      legend columns=-1,
      /tikz/every even column/.append style={
        column sep=1em
      },
    },
  },
  qwen ott/.style={
    qwenott,
    line width=1pt,
    solid,
    mark=none,
  },
  qwen numina/.style={
    qwennumina,
    line width=1pt,
    solid,
    mark=none,
  },
  gemma math/.style={
    gemmamath,
    line width=1pt,
    solid,
    mark=none,
  },
  gemma numina/.style={
    gemmanumina,
    line width=1pt,
    solid,
    mark=none,
  },
}

\tikzset{
  correlation box/.style={
    font=\fontsize{6}{7.5}\selectfont,
    align=left,
    anchor=north west,
    fill=white,
    fill opacity=0.90,
    text opacity=1,
    draw=black!20,
    inner xsep=1pt,
    inner ysep=0.8pt,
    rounded corners=0.8pt,
  },
  panel label/.style={
    anchor=north west,
    font=\small\bfseries,
    inner sep=1pt,
  },
}

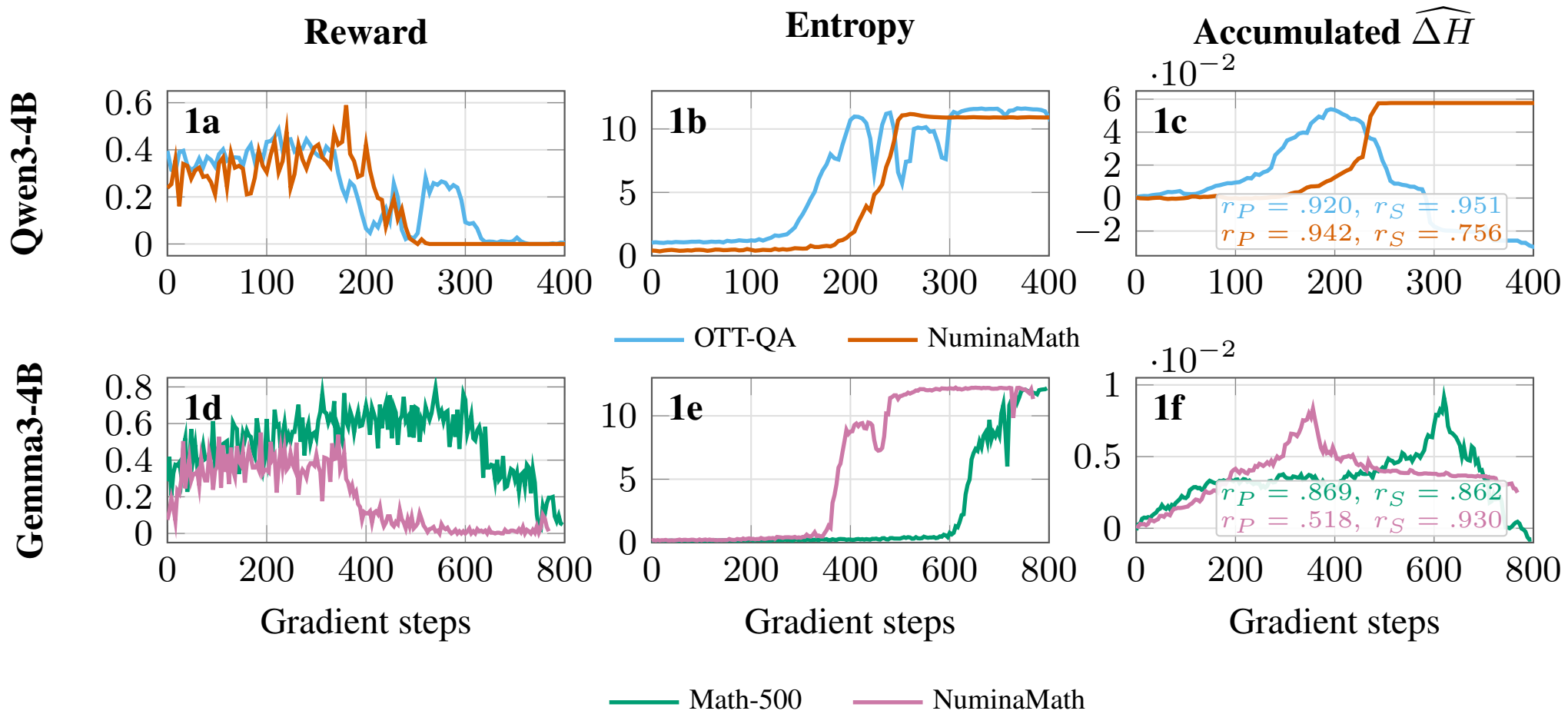
\begin{figure*}[t]
  \centering

  \begin{subcaptiongroup}


    \phantomsubcaption
    \label{fig:entropy-collapse:qwen-reward}

    \phantomsubcaption
    \label{fig:entropy-collapse:qwen-entropy}

    \phantomsubcaption
    \label{fig:entropy-collapse:qwen-deltah}

    \phantomsubcaption
    \label{fig:entropy-collapse:gemma-reward}

    \phantomsubcaption
    \label{fig:entropy-collapse:gemma-entropy}

    \phantomsubcaption
    \label{fig:entropy-collapse:gemma-deltah}

    \begin{adjustbox}{max width=\textwidth}
    \begin{tikzpicture}

      \begin{groupplot}[
        plotbase,
        group style={
          group size=3 by 2,
          horizontal sep=0.75cm,
          vertical sep=1.05cm,
        },
      ]


        \nextgroupplot[
          xmin=0,
          xmax=400,
          ymin=-0.05,
          ymax=0.65,
          legend to name=qwenlegend,
        ]
          \addplot[qwen ott]
            table[
              x=gradient_step,
              y=reward,
              col sep=comma
            ]
            {plots/entropy-first-order-collapse/data/qwen-ott.csv};
          \addlegendentry{OTT-QA}

          \addplot[qwen numina]
            table[
              x=gradient_step,
              y=reward,
              col sep=comma
            ]
            {plots/entropy-first-order-collapse/data/qwen-numina.csv};
          \addlegendentry{NuminaMath}


        \nextgroupplot[
          xmin=0,
          xmax=400,
          ymin=0,
          ymax=13,
        ]
          \addplot[qwen ott]
            table[
              x=gradient_step,
              y=entropy,
              col sep=comma
            ]
            {plots/entropy-first-order-collapse/data/qwen-ott.csv};

          \addplot[qwen numina]
            table[
              x=gradient_step,
              y=entropy,
              col sep=comma
            ]
            {plots/entropy-first-order-collapse/data/qwen-numina.csv};


        \nextgroupplot[
          xmin=0,
          xmax=400,
          ymin=-0.035,
          ymax=0.065,
          scaled y ticks=true,
        ]
          \addplot[qwen ott]
            table[
              x=gradient_step,
              y=cumulative_first_order_entropy_change,
              col sep=comma
            ]
            {plots/entropy-first-order-collapse/data/qwen-ott.csv};

          \addplot[qwen numina]
            table[
              x=gradient_step,
              y=cumulative_first_order_entropy_change,
              col sep=comma
            ]
            {plots/entropy-first-order-collapse/data/qwen-numina.csv};

          \node[correlation box]
            at (rel axis cs:0.20,0.38) {
              \textcolor{qwenott}
                {$r_P=.920,\ r_S=.951$}\\[-0.3ex]
              \textcolor{qwennumina}
                {$r_P=.942,\ r_S=.756$}
            };


        \nextgroupplot[
          xmin=0,
          xmax=800,
          ymin=-0.05,
          ymax=0.85,
          xlabel={Gradient steps},
          legend to name=gemmalegend,
        ]
          \addplot[gemma math]
            table[
              x=gradient_step,
              y=reward,
              col sep=comma
            ]
            {plots/entropy-first-order-collapse/data/gemma-math.csv};
          \addlegendentry{Math-500}

          \addplot[gemma numina]
            table[
              x=gradient_step,
              y=reward,
              col sep=comma
            ]
            {plots/entropy-first-order-collapse/data/gemma-numina.csv};
          \addlegendentry{NuminaMath}


        \nextgroupplot[
          xmin=0,
          xmax=800,
          ymin=0,
          ymax=13,
          xlabel={Gradient steps},
        ]
          \addplot[gemma math]
            table[
              x=gradient_step,
              y=entropy,
              col sep=comma
            ]
            {plots/entropy-first-order-collapse/data/gemma-math.csv};

          \addplot[gemma numina]
            table[
              x=gradient_step,
              y=entropy,
              col sep=comma
            ]
            {plots/entropy-first-order-collapse/data/gemma-numina.csv};


        \nextgroupplot[
          xmin=0,
          xmax=800,
          ymin=-0.001,
          ymax=0.0105,
          xlabel={Gradient steps},
          scaled y ticks=true,
        ]
          \addplot[gemma math]
            table[
              x=gradient_step,
              y=cumulative_first_order_entropy_change,
              col sep=comma
            ]
            {plots/entropy-first-order-collapse/data/gemma-math.csv};

          \addplot[gemma numina]
            table[
              x=gradient_step,
              y=cumulative_first_order_entropy_change,
              col sep=comma
            ]
            {plots/entropy-first-order-collapse/data/gemma-numina.csv};

          \node[correlation box]
            at (rel axis cs:0.20,0.38) {
              \textcolor{gemmamath}
                {$r_P=.869,\ r_S=.862$}\\[-0.3ex]
              \textcolor{gemmanumina}
                {$r_P=.518,\ r_S=.930$}
            };

      \end{groupplot}


      \node[panel label]
        at ([xshift=1mm,yshift=-1mm]group c1r1.north west)
        {\ref{fig:entropy-collapse:qwen-reward}};

      \node[panel label]
        at ([xshift=1mm,yshift=-1mm]group c2r1.north west)
        {\ref{fig:entropy-collapse:qwen-entropy}};

      \node[panel label]
        at ([xshift=1mm,yshift=-1mm]group c3r1.north west)
        {\ref{fig:entropy-collapse:qwen-deltah}};

      \node[panel label]
        at ([xshift=1mm,yshift=-1mm]group c1r2.north west)
        {\ref{fig:entropy-collapse:gemma-reward}};

      \node[panel label]
        at ([xshift=1mm,yshift=-1mm]group c2r2.north west)
        {\ref{fig:entropy-collapse:gemma-entropy}};

      \node[panel label]
        at ([xshift=1mm,yshift=-1mm]group c3r2.north west)
        {\ref{fig:entropy-collapse:gemma-deltah}};


      \node[
        font=\small\bfseries,
        anchor=south
      ] at ([yshift=0.28cm]group c1r1.north)
        {Reward};

      \node[
        font=\small\bfseries,
        anchor=south
      ] at ([yshift=0.28cm]group c2r1.north)
        {Entropy};

      \node[
        font=\small\bfseries,
        anchor=south
      ] at ([yshift=0.28cm]group c3r1.north)
        {Accumulated $\widehat{\Delta H}$};


      \node[
        rotate=90,
        font=\small\bfseries,
        anchor=south
      ] at ([xshift=-0.95cm]group c1r1.west)
        {Qwen3-4B};

      \node[
        rotate=90,
        font=\small\bfseries,
        anchor=south
      ] at ([xshift=-0.95cm]group c1r2.west)
        {Gemma3-4B};


      \path (group c1r1.south west) --
        node[
          anchor=north,
          yshift=-3.5mm
        ] {
          \pgfplotslegendfromname{qwenlegend}
        }
        (group c3r1.south east);

      \path (group c1r2.south west) --
        node[
          anchor=north,
          yshift=-10mm
        ] {
          \pgfplotslegendfromname{gemmalegend}
        }
        (group c3r2.south east);

    \end{tikzpicture}
    \end{adjustbox}

  \end{subcaptiongroup}

  \vspace{-0.2cm}

  \caption{
    Each row shows one model family across two datasets. In both cases,
    reward collapse
    (\ref{fig:entropy-collapse:qwen-reward},
    \ref{fig:entropy-collapse:gemma-reward})
    coincides with a sharp increase in policy entropy
    (\ref{fig:entropy-collapse:qwen-entropy},
    \ref{fig:entropy-collapse:gemma-entropy}).
    We sum $\widehat{\Delta H}$ over tokens at each gradient step and
    accumulate it over training
    (\ref{fig:entropy-collapse:qwen-deltah},
    \ref{fig:entropy-collapse:gemma-deltah}).
    Before collapse, accumulated $\widehat{\Delta H}$ closely follows
    the observed entropy dynamics. Annotations report Pearson ($r_P$)
    and Spearman ($r_S$) correlations between accumulated
    $\widehat{\Delta H}$ and policy entropy up to the collapse step.
  }
  \label{fig:entropy-first-order-collapse}
  \vspace{-4mm}
\end{figure*}

\paragraph{Entropy growth accompanies collapse.}

For a next-token distribution $p$, entropy
$H(p)=-\sum_j p_j\log p_j$ measures how concentrated the model's
prediction is over the vocabulary. An increase in entropy corresponds to
probability mass becoming less concentrated among likely next tokens, and
therefore provides a direct measure of increasing uncertainty in the
trainer's predictions.

Fig~\ref{fig:entropy-first-order-collapse} compares reward (Fig.~\ref{fig:entropy-collapse:qwen-reward},~\ref{fig:entropy-collapse:gemma-reward}) and entropy (Fig.~\ref{fig:entropy-collapse:qwen-entropy},~\ref{fig:entropy-collapse:gemma-entropy}) during collapse-prone training runs. We measure entropy under the current trainer policy $\pi_\theta$, while evaluating it on contexts generated by the
rollout policy: for prompts $q\sim\mathcal{D}$ and rollouts
$x\sim\pi_{\mathrm{roll}}(\cdot\mid q)$, we average
$H(\pi_\theta(\cdot\mid q,x_{<t}))$ over token positions and sampled
rollouts. Across both models and datasets, the onset of reward collapse is
accompanied by a sharp increase in this quantity. We define collapse at step $t$ if the reward falls below $50\%$ of its value
at $t$ within the following 50 gradient steps and remains below this level
for at least 10 consecutive steps. This occurs around step
160 for Qwen3-4B on OTT-QA, 200 for Qwen3-4B on NuminaMath, 360 for
Gemma3-4B on NuminaMath, and 600 for Gemma3-4B on Math-500
(Fig~\ref{fig:entropy-collapse:qwen-reward}, ~\ref{fig:entropy-collapse:gemma-reward}). 

\paragraph{First-order entropy change tracks training entropy.}

We next relate the observed entropy increase to the gradient contributed by
individual tokens. Let $p$ denote the trainer's next-token distribution and
$p_s$ the probability assigned to the sampled token. For an active,
unclipped policy-gradient update with advantage $A$ and importance ratio
$\rho_s$, the first-order entropy change in logit space is
\begin{equation}
\label{eq:delta_entropy}
\widehat{\Delta H}
=
-\eta \rho_s A
\left[
p_s\log p_s
-
\sum_j p_j^2\log p_j
+
H(p)
\left(
p_s-\sum_{j=1}^{\mathcal{V}} p_j^2
\right)
\right],
\end{equation}
where $\rho_s =
\pi_\theta(s\mid q,x_{<t})/
\pi_{\mathrm{old}}(s\mid q,x_{<t})$. We follow the derivation in \citet{yss2026gradient, wang2026entropy}, detailed proof is provided in App~\ref{app:first-order-entropy}. 

Eq~\ref{eq:delta_entropy} is a local first-order estimate rather than the exact entropy change of the full network. The derivation treats the logits at a context locally: in the neural network, parameters are shared across tokens and contexts, so a minibatch update introduces cross-token interactions that are absent from above calculation. It also holds the sampled context distribution fixed, and therefore does not account for how the update changes the generated contexts from future rollouts. We use $\widehat{\Delta H}$ as a tractable estimate of the direction of the entropy change rather than its exact magnitude.

Despite these approximations, accumulated $\widehat{\Delta H}$ follows the same trend as the observed entropy prior to collapse. We sum $\widehat{\Delta H}$ over tokens at each gradient step and accumulate it during training (see App~\ref{app:accumulated_entropy_change}). Across the four settings, the resulting quantity has Spearman correlations and Pearson correlations ranging from $0.756$ to $0.951$ and $0.518$ to $0.942$ respectively with the measured entropy prior to collapse (Fig.~\ref{fig:entropy-collapse:qwen-deltah}, ~\ref{fig:entropy-collapse:gemma-deltah}). In particular, cumulative $\widehat{\Delta H}$ begins to increase in the same region where trainer entropy begins to rise.

Eq.~\ref{eq:delta_entropy} also gives a direct way to intervene on the updates. In Section~\ref{sec:experiments}, we show that suppressing tokens with large positive predicted entropy change consistently improves stability in collapse-prone settings. In contrast, suppressing tokens with large negative predicted entropy change accelerates collapse. These interventions provide evidence that the sign of the first-order entropy change is not only correlated with instability, but can also be used to control it.

\section{Experiments}
\label{sec:experiments}

%
%

\begin{figure*}[t]
\centering

\begin{subcaptiongroup}

\phantomsubcaption\label{fig:main-training-results:ott-qwen}
\phantomsubcaption\label{fig:main-training-results:numina-qwen}
\phantomsubcaption\label{fig:main-training-results:math-gemma}
\phantomsubcaption\label{fig:main-training-results:numina-gemma}

\definecolor{FilterBlue}{HTML}{0072B2}
\definecolor{FilterOrange}{HTML}{E69F00}
\definecolor{FilterRed}{HTML}{D55E00}
\definecolor{FilterGray}{HTML}{6B7280}

\providecommand{\EntropyFilteringDataDir}{plots/entropy-filtering-reward-entropy/data}

\pgfmathsetlengthmacro{\EntropyFilteringPanelWidth}{0.30\textwidth}
\pgfmathsetlengthmacro{\EntropyFilteringPanelHeight}{0.22\textwidth}

\pgfplotsset{
  entropy filter baseline/.style={FilterBlue, very thick, solid, no marks},
  entropy filter upper/.style={FilterOrange, very thick, solid, no marks},
  entropy filter two sided/.style={FilterRed, very thick, solid, no marks},
}

\newcommand{\AddRewardCurves}[1]{%
  \addplot[entropy filter baseline]
    table[x=gradient_step, y=reward, col sep=comma]
    {\EntropyFilteringDataDir/#1-baseline.csv};
  \addplot[entropy filter upper]
    table[x=gradient_step, y=reward, col sep=comma]
    {\EntropyFilteringDataDir/#1-upper.csv};
  \addplot[entropy filter two sided]
    table[x=gradient_step, y=reward, col sep=comma]
    {\EntropyFilteringDataDir/#1-two-sided.csv};
}

\newcommand{\AddEntropyCurves}[1]{%
  \addplot[entropy filter baseline]
    table[x=gradient_step, y=entropy, col sep=comma]
    {\EntropyFilteringDataDir/#1-baseline.csv};
  \addplot[entropy filter upper]
    table[x=gradient_step, y=entropy, col sep=comma]
    {\EntropyFilteringDataDir/#1-upper.csv};
  \addplot[entropy filter two sided]
    table[x=gradient_step, y=entropy, col sep=comma]
    {\EntropyFilteringDataDir/#1-two-sided.csv};
}

\begin{tikzpicture}
  \begin{groupplot}[
    group style={
      group size=4 by 2,
      horizontal sep=0.65cm,
      vertical sep=2.0cm,
    },
    width=\EntropyFilteringPanelWidth,
    height=\EntropyFilteringPanelHeight,
    xmin=0,
    unbounded coords=jump,
    grid=major,
    grid style={FilterGray!18},
    axis line style={FilterGray!70},
    tick style={FilterGray!70},
    tick label style={font=\scriptsize},
    label style={font=\small},
    title style={font=\small, yshift=-0.5ex},
    scaled y ticks=false,
    yticklabel style={
      font=\scriptsize,
      /pgf/number format/fixed,
      /pgf/number format/precision=2,
      /pgf/number format/skip 0.=false,
    },
    legend style={
      draw=none,
      font=\small,
      legend columns=-1,
      /tikz/every even column/.append style={column sep=1.0em},
    },
  ]
    \nextgroupplot[
      title={Reward}, xlabel={Gradient steps},
      xmax=400, xtick={0,100,200,300,400,500},
      ymin=0, ymax=0.65, ytick={0,0.2,0.4,0.6,0.8},
      legend to name=entropyfilterlegend,
    ]
      \AddRewardCurves{ott-qwen}
      \addlegendentry{Baseline}
      \addlegendentry{$\widehat{\Delta H}<\tau^{+}$}
      \addlegendentry{$\tau^{-}<\widehat{\Delta H}<\tau^{+}$}

    \nextgroupplot[
      title={Entropy}, xlabel={Gradient steps},
      xmax=400, xtick={0,100,200,300,400,500},
      ymin=0, ymax=12.5, ytick={0,3,6,9,12},
    ]
      \AddEntropyCurves{ott-qwen}

    \nextgroupplot[
      title={Reward}, xlabel={Gradient steps},
      xmax=400, xtick={0,100,200,300,400,500},
      ymin=0, ymax=0.70, ytick={0,0.2,0.4,0.6,0.8},
    ]
      \AddRewardCurves{numina-qwen}

    \nextgroupplot[
      title={Entropy}, xlabel={Gradient steps},
      xmax=400, xtick={0,100,200,300,400,500},
      ymin=0, ymax=12.5, ytick={0,3,6,9,12},
    ]
      \AddEntropyCurves{numina-qwen}

    \nextgroupplot[
      title={Reward}, xlabel={Gradient steps},
      xmax=1000, xtick={0,250,500,750,1000},
      ymin=0, ymax=0.90, ytick={0,0.2,0.4,0.6,0.8},
    ]
      \AddRewardCurves{math-gemma}

    \nextgroupplot[
      title={Entropy}, xlabel={Gradient steps},
      xmax=1000, xtick={0,250,500,750,1000},
      ymin=0, ymax=13.0, ytick={0,3,6,9,12},
    ]
      \AddEntropyCurves{math-gemma}

    \nextgroupplot[
      title={Reward}, xlabel={Gradient steps},
      xmax=600, xtick={0,200,400,600,800},
      ymin=0, ymax=0.62, ytick={0,0.2,0.4,0.6,0.8},
    ]
      \AddRewardCurves{numina-gemma}

    \nextgroupplot[
      title={Entropy}, xlabel={Gradient steps},
      xmax=600, xtick={0,200,400,600,800},
      ymin=0, ymax=13.0, ytick={0,3,6,9,12},
    ]
      \AddEntropyCurves{numina-gemma}
  \end{groupplot}

\path (group c1r1.north west) --
  node[
    anchor=south,
    yshift=5mm,
    font=\small\bfseries
  ] {
    (\subref{fig:main-training-results:ott-qwen})
    OTT-QA--Qwen
  }
  (group c2r1.north east);

\path (group c3r1.north west) --
  node[
    anchor=south,
    yshift=5mm,
    font=\small\bfseries
  ] {
    (\subref{fig:main-training-results:numina-qwen})
    Numina--Qwen
  }
  (group c4r1.north east);

\path (group c1r2.north west) --
  node[
    anchor=south,
    yshift=5mm,
    font=\small\bfseries
  ] {
    (\subref{fig:main-training-results:math-gemma})
    MATH--Gemma
  }
  (group c2r2.north east);

\path (group c3r2.north west) --
  node[
    anchor=south,
    yshift=5mm,
    font=\small\bfseries
  ] {
    (\subref{fig:main-training-results:numina-gemma})
    Numina--Gemma
  }
  (group c4r2.north east);

\path (group c1r2.south west) --
  node[
    anchor=north,
    yshift=-9mm
  ] {
    \pgfplotslegendfromname{entropyfilterlegend}
  }
  (group c4r2.south east);
\end{tikzpicture}

\end{subcaptiongroup}

\vspace{-4mm}
\caption{Here we show reward and policy entropy for the baseline, an upper entropy-change filter $\widehat{\Delta H}<\tau^+$, and a two-sided filter that retains updates with $\tau^-<\widehat{\Delta H}<\tau^+$. We set $\tau^+=0.1$ for Qwen3-4B on OTT-QA and $\tau^+=0.2$ for the remaining settings, with $\tau^-=-0.1$ throughout. Across all four model--dataset settings, filtering entropy-increasing updates stabilizes training, whereas additionally filtering entropy-decreasing updates leads to earlier entropy growth and reward collapse. Collapsed runs are truncated at point of collapse.}
\label{fig:main-training-results}
\vspace{-4mm}
\end{figure*}
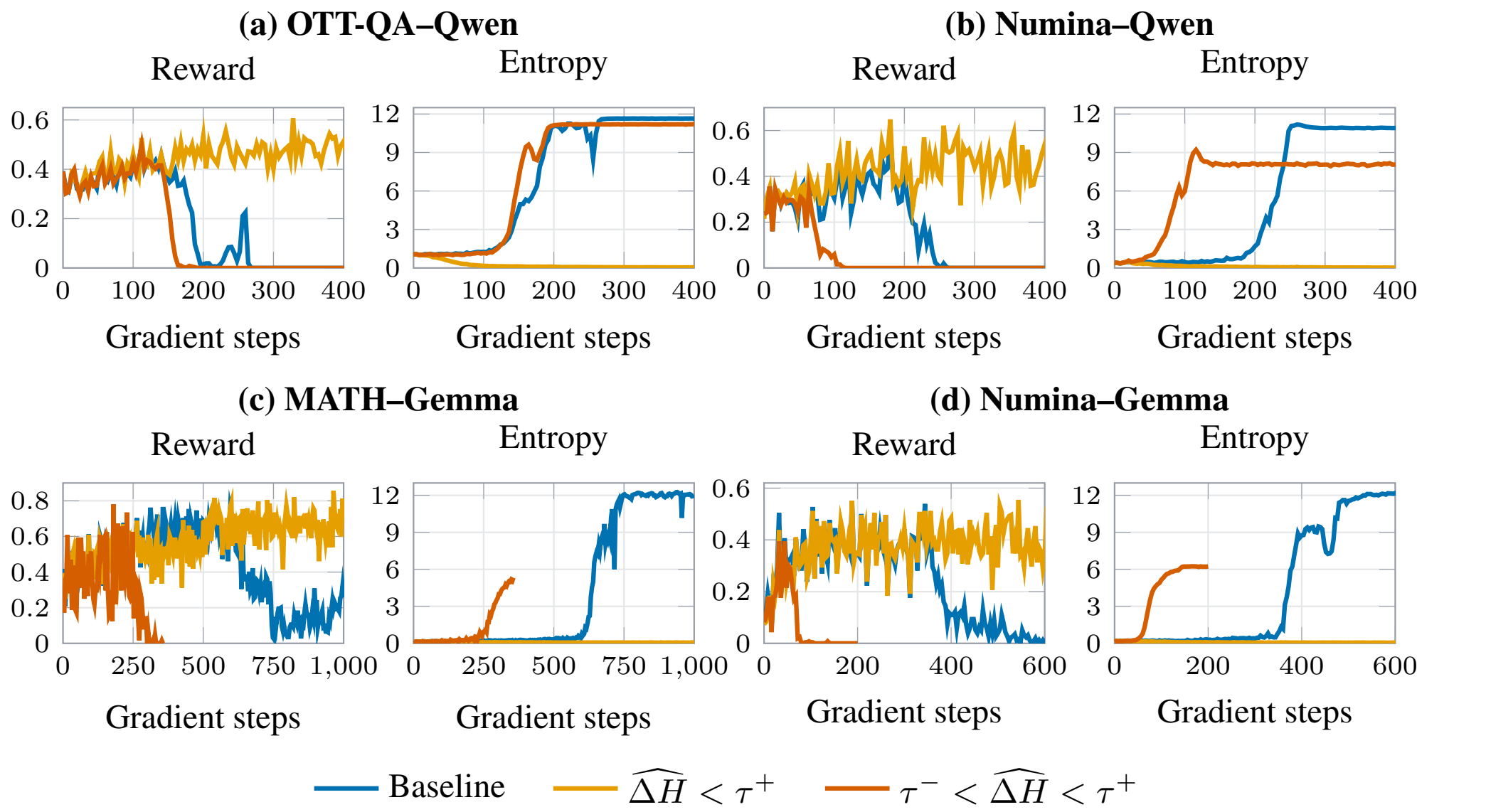

\paragraph{Main Results.}\label{sec:entropy-thresholding}
We evaluate $\widehat{\Delta H}$-based filtering on three datasets:
OTT-QA~\cite{chen2020open}, MATH-500~\cite{lightman2024let}, and
NuminaMath-LEAN \cite{wang2025kimina} (App.~\ref{app:exp-setup}), using Qwen3-4B~\cite{yang2025qwen3}
and Gemma3-4B~\cite{gemmateam2025gemma3technicalreport}
(Fig.~\ref{fig:main-training-results}). All experiments use four policy updates per rollout batch, introducing policy lag across successive updates. Full training details are provided in App.~\ref{app:exp-setup}. For each
model--dataset pair, we select a DAPO~\citep{yu2026dapo} configuration that exhibits collapse (see App. ~\ref{app:exp-setup})
and compare the baseline against two filters: an upper filter that removes
updates with $\widehat{\Delta H}>\tau^+$, and a two-sided filter that retains
only updates satisfying $\tau^-<\widehat{\Delta H}<\tau^+$. We omit learning-rate($\eta$) from $\widehat{\Delta H}$ in these experiments.

Previous importance-ratio clipping results do not show a consistent relation
between clipping strength and stability: tighter clipping is not uniformly
more stable than weaker clipping~\citep{yss2026gradient}. In contrast,
$\widehat{\Delta H}$-based filtering produces a consistent directional effect
across all four model--dataset combinations in
Fig.~\ref{fig:main-training-results}. In each case, baseline collapse coincides with a sharp increase in policy
entropy. Removing entropy-increasing updates keeps entropy controlled and
remains stable over the plotted training horizon in all four settings.
Conversely, additionally removing entropy-decreasing updates causes an
earlier rise in entropy and earlier collapse than the baseline in all four
settings. This directional effect is consistent across both models, all three
datasets, and the two values of $\tau^+$ used in our experiments. Additional experiments in App.~\ref{app:additional-exps} further examine the effect of filtering thresholds on OTT-QA with Qwen3-4B. Consistent with the main results,
decreasing $\tau^+$ removes more entropy-increasing updates and delays collapse, while increasing $\tau^{-}$ removes more entropy-decreasing updates and accelerates collapse (Fig.~\ref{fig:threshold-seeds:42-reward},~\ref{fig:threshold-seeds:42-entropy},~\ref{fig:threshold-seeds:42-deltah}).


The contrast is especially pronounced for Gemma3-4B on MATH-500
(Fig.~\ref{fig:main-training-results:math-gemma}). The baseline begins to collapse at roughly
500 gradient steps. The run filtering updates with
$\widehat{\Delta H}>0.2$ remains stable over the training horizon, whereas
the two-sided filter retaining only
$-0.1 < \widehat{\Delta H} < 0.2$ collapses at approximately 250 steps.
This shows that the stability improvement from filtering entropy-increasing
updates cannot be explained solely by reducing the overall amount of optimization signal. Removing additional entropy-decreasing updates instead accelerates collapse. We report hyper-parameter sweep and multi-seed results for OTT-QA with Qwen3-4B (Fig.~\ref{fig:ott-qwen-threshold-seeds}) and cumulative $\widehat{\Delta H}$ trajectories for all configurations (Fig.~\ref{fig:entropy-filtering-seed42}) in
Appendix~\ref{app:additional-exps}.

\paragraph{Revisiting Existing Methods.}
\label{sec:existing-methods}
%
%

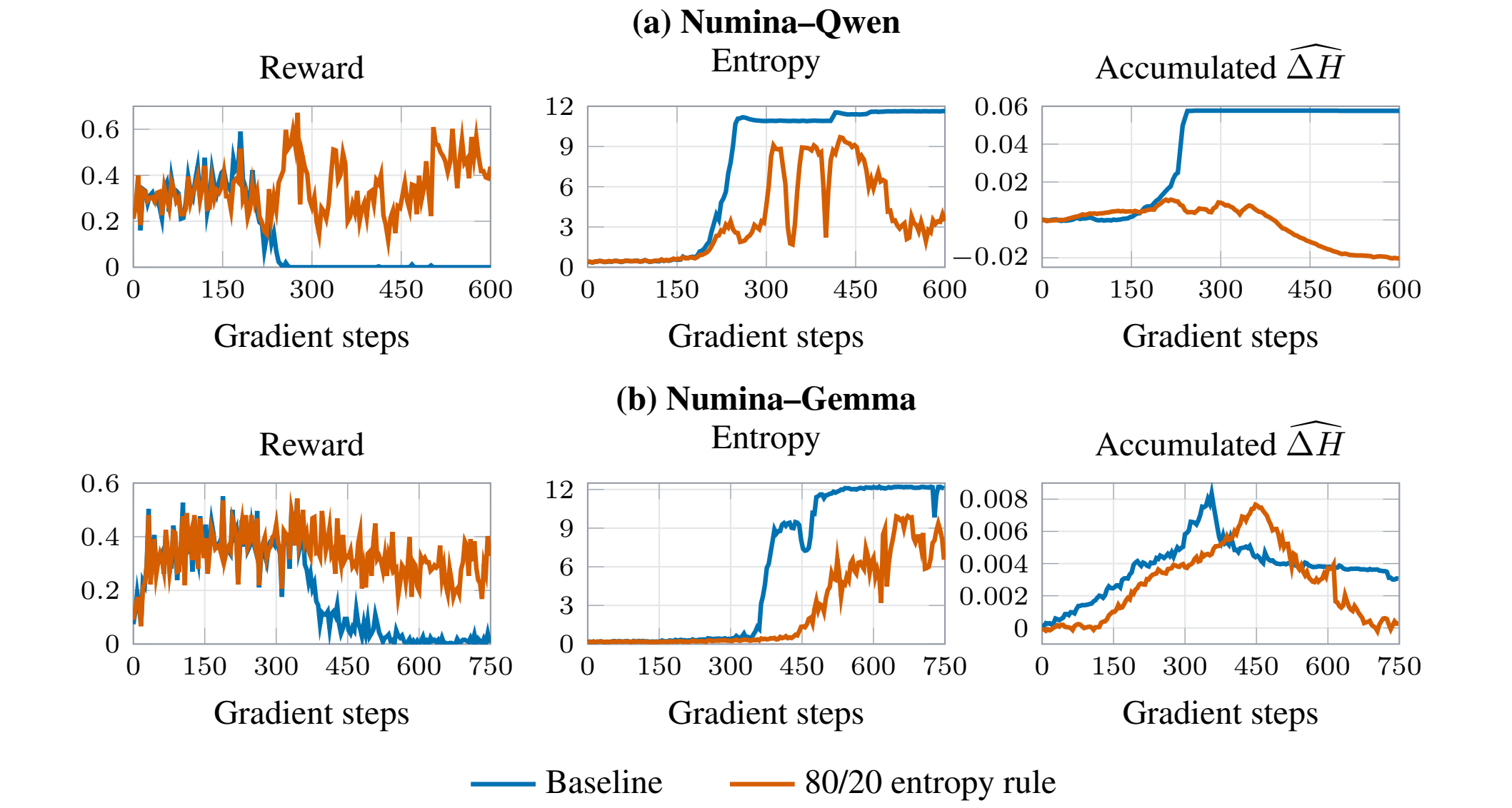
\begin{figure*}[t]
\centering
\vspace{-4mm}

\definecolor{EntropyRuleBlue}{HTML}{0072B2}
\definecolor{EntropyRuleRed}{HTML}{D55E00}
\definecolor{EntropyRuleGray}{HTML}{6B7280}

\providecommand{\EntropyRuleDataDir}{plots/entropy-80-20-rule/data}

\pgfmathsetlengthmacro{\EntropyRulePanelWidth}{0.35\textwidth}
\pgfmathsetlengthmacro{\EntropyRulePanelHeight}{0.22\textwidth}

\begin{subcaptiongroup}

\phantomsubcaption
\label{fig:entropy-80-20-rule:numina-qwen}

\phantomsubcaption
\label{fig:entropy-80-20-rule:numina-gemma}

\begin{tikzpicture}
  \begin{groupplot}[
    group style={
      group size=3 by 2,
      horizontal sep=0.9cm,
      vertical sep=2.0cm,
    },
    width=\EntropyRulePanelWidth,
    height=\EntropyRulePanelHeight,
    xmin=0,
    unbounded coords=jump,
    grid=major,
    grid style={EntropyRuleGray!18},
    axis line style={EntropyRuleGray!70},
    tick style={EntropyRuleGray!70},
    tick label style={font=\scriptsize},
    label style={font=\small},
    title style={font=\small, yshift=-0.5ex},
    scaled y ticks=false,
    yticklabel style={
      font=\scriptsize,
      /pgf/number format/fixed,
      /pgf/number format/precision=3,
      /pgf/number format/skip 0.=false,
    },
    legend style={
      draw=none,
      fill=none,
      font=\small,
      legend columns=-1,
      cells={anchor=west},
      /tikz/every even column/.append style={
        column sep=1.5em
      },
    },
  ]
    \nextgroupplot[
      title={Reward},
      xmax=600,
      xtick={0,150,300,450,600},
      xlabel={Gradient steps},
      ymin=0,
      ymax=0.7,
      ytick={0,0.2,0.4,0.6},
      legend to name=entropyrulelegend,
    ]
      \addplot[EntropyRuleBlue, very thick, solid, no marks]
        table[x=gradient_step, y=reward, col sep=comma]
        {\EntropyRuleDataDir/numina-qwen-baseline.csv};
      \addlegendentry{Baseline}
      \addplot[EntropyRuleRed, very thick, solid, no marks]
        table[x=gradient_step, y=reward, col sep=comma]
        {\EntropyRuleDataDir/numina-qwen-80-20.csv};
      \addlegendentry{80/20 entropy rule}

    \nextgroupplot[
      title={Entropy},
      xmax=600,
      xtick={0,150,300,450,600},
      xlabel={Gradient steps},
      ymin=0,
      ymax=12,
      ytick={0,3,6,9,12},
    ]
      \addplot[EntropyRuleBlue, very thick, solid, no marks]
        table[x=gradient_step, y=entropy, col sep=comma]
        {\EntropyRuleDataDir/numina-qwen-baseline.csv};
      \addplot[EntropyRuleRed, very thick, solid, no marks]
        table[x=gradient_step, y=entropy, col sep=comma]
        {\EntropyRuleDataDir/numina-qwen-80-20.csv};

    \nextgroupplot[
      title={Accumulated $\widehat{\Delta H}$},
      xmax=600,
      xtick={0,150,300,450,600},
      xlabel={Gradient steps},
      ymin=-0.025,
      ymax=0.06,
      ytick={-0.02,0,0.02,0.04,0.06},
    ]
      \addplot[EntropyRuleBlue, very thick, solid, no marks]
        table[x=gradient_step, y=cumulative_first_order_entropy_change, col sep=comma]
        {\EntropyRuleDataDir/numina-qwen-baseline.csv};
      \addplot[EntropyRuleRed, very thick, solid, no marks]
        table[x=gradient_step, y=cumulative_first_order_entropy_change, col sep=comma]
        {\EntropyRuleDataDir/numina-qwen-80-20.csv};

    \nextgroupplot[
      title={Reward},
      xmax=750,
      xtick={0,150,300,450,600,750},
      xlabel={Gradient steps},
      ymin=0,
      ymax=0.6,
      ytick={0,0.2,0.4,0.6},
    ]
      \addplot[EntropyRuleBlue, very thick, solid, no marks]
        table[x=gradient_step, y=reward, col sep=comma]
        {\EntropyRuleDataDir/numina-gemma-baseline.csv};
      \addplot[EntropyRuleRed, very thick, solid, no marks]
        table[x=gradient_step, y=reward, col sep=comma]
        {\EntropyRuleDataDir/numina-gemma-80-20.csv};

    \nextgroupplot[
      title={Entropy},
      xmax=750,
      xtick={0,150,300,450,600,750},
      xlabel={Gradient steps},
      ymin=0,
      ymax=12.5,
      ytick={0,3,6,9,12},
    ]
      \addplot[EntropyRuleBlue, very thick, solid, no marks]
        table[x=gradient_step, y=entropy, col sep=comma]
        {\EntropyRuleDataDir/numina-gemma-baseline.csv};
      \addplot[EntropyRuleRed, very thick, solid, no marks]
        table[x=gradient_step, y=entropy, col sep=comma]
        {\EntropyRuleDataDir/numina-gemma-80-20.csv};

    \nextgroupplot[
      title={Accumulated $\widehat{\Delta H}$},
      xmax=750,
      xtick={0,150,300,450,600,750},
      xlabel={Gradient steps},
      ymin=-0.001,
      ymax=0.009,
      ytick={0,0.002,0.004,0.006,0.008},
    ]
      \addplot[EntropyRuleBlue, very thick, solid, no marks]
        table[x=gradient_step, y=cumulative_first_order_entropy_change, col sep=comma]
        {\EntropyRuleDataDir/numina-gemma-baseline.csv};
      \addplot[EntropyRuleRed, very thick, solid, no marks]
        table[x=gradient_step, y=cumulative_first_order_entropy_change, col sep=comma]
        {\EntropyRuleDataDir/numina-gemma-80-20.csv};
  \end{groupplot}

\path (group c1r1.north west) --
  node[
    anchor=south,
    yshift=5mm,
    font=\small\bfseries
  ] {
    (\subref{fig:entropy-80-20-rule:numina-qwen})
    Numina--Qwen
  }
  (group c3r1.north east);

\path (group c1r2.north west) --
  node[
    anchor=south,
    yshift=5mm,
    font=\small\bfseries
  ] {
    (\subref{fig:entropy-80-20-rule:numina-gemma})
    Numina--Gemma
  }
  (group c3r2.north east);
\path (group c1r2.south west) --
  node[
    anchor=north,
    yshift=-9mm
  ] {
    \pgfplotslegendfromname{entropyrulelegend}
  }
  (group c3r2.south east);

\end{tikzpicture}

\end{subcaptiongroup}
\vspace{-3mm}
\caption{Reward, policy entropy, and accumulated $\widehat{\Delta H}$ of baseline and the 80/20 entropy rule~\cite{wang2026beyond}. Training on high entropy tokens only substantially improves stability for Numina--Qwen, whereas the improvement is weaker for Numina--Gemma. Across both settings, stronger stabilization is accompanied by a larger reduction in accumulated $\widehat{\Delta H}$ and lower subsequent entropy growth.}
\vspace{-6mm}
\label{fig:entropy-80-20-rule}
\end{figure*}

We next examine existing stabilization methods through the first-order
entropy-change perspective developed above. We ask whether their effect on
stability coincides with their effect on entropy growth. \cite{wang2026beyond} retain only the highest-entropy $20\%$ of token updates. In Fig.~\ref{fig:entropy-80-20-rule}, this filter
substantially reduces accumulated $\widehat{\Delta H}$ for Numina-Qwen,
keeps entropy lower, and avoids the complete reward collapse of the baseline.
For Numina-Gemma, the stabilizing effect is weaker: entropy growth is delayed but not
eliminated, and reward still degrades later in training. Thus, the degree of
stabilization tracks how strongly the method controls entropy growth.

%
%

\begin{figure*}[t]
\centering

\definecolor{PeakColor}{HTML}{0072B2}
\definecolor{ValleyColor}{HTML}{D55E00}
\definecolor{PeakValleyGray}{HTML}{6B7280}

\providecommand{\PeakValleyDataDir}
  {plots/peak-valley-entropy/data}

\pgfmathsetlengthmacro{\PeakValleyPanelWidth}{0.38\textwidth}
\pgfmathsetlengthmacro{\PeakValleyPanelHeight}{0.25\textwidth}

\begin{tikzpicture}
  \begin{groupplot}[
    group style={
      group size=3 by 1,
      horizontal sep=0.9cm,
    },
    width=\PeakValleyPanelWidth,
    height=\PeakValleyPanelHeight,
    xmin=0,
    xmax=800,
    xtick={0,200,400,600,800},
    xlabel={Gradient steps},
    unbounded coords=jump,
    grid=major,
    grid style={PeakValleyGray!18},
    axis line style={PeakValleyGray!70},
    tick style={PeakValleyGray!70},
    tick label style={font=\scriptsize},
    label style={font=\small},
    title style={
      font=\small,
      yshift=-0.5ex
    },
    scaled y ticks=false,
    yticklabel style={
      font=\scriptsize,
      /pgf/number format/fixed,
      /pgf/number format/precision=3,
      /pgf/number format/skip 0.=false,
    },
    legend style={
      draw=none,
      fill=none,
      font=\small,
      legend columns=-1,
      cells={anchor=west},
      /tikz/every even column/.append style={
        column sep=1.5em
      },
    },
  ]

    \nextgroupplot[
      title={Reward},
      ymin=0,
      ymax=0.65,
      ytick={0,0.2,0.4,0.6},
      legend to name=peakvalleylegend,
    ]
      \addplot[
        PeakColor,
        very thick,
        solid,
        no marks
      ]
        table[
          x=gradient_step,
          y=reward,
          col sep=comma
        ]
        {\PeakValleyDataDir/peak.csv};
      \addlegendentry{Peak tokens $\left(p_s>\sum_{j=1}^{\mathcal{V}} p_j^2\right)$}

      \addplot[
        ValleyColor,
        very thick,
        solid,
        no marks
      ]
        table[
          x=gradient_step,
          y=reward,
          col sep=comma
        ]
        {\PeakValleyDataDir/valley.csv};
      \addlegendentry{Valley tokens $\left(p_s<=\sum_{j=1}^{\mathcal{V}} p_j^2\right)$}

    \nextgroupplot[
      title={Entropy},
      ymin=0,
      ymax=12,
      ytick={0,3,6,9,12},
    ]
      \addplot[
        PeakColor,
        very thick,
        solid,
        no marks
      ]
        table[
          x=gradient_step,
          y=entropy,
          col sep=comma
        ]
        {\PeakValleyDataDir/peak.csv};

      \addplot[
        ValleyColor,
        very thick,
        solid,
        no marks
      ]
        table[
          x=gradient_step,
          y=entropy,
          col sep=comma
        ]
        {\PeakValleyDataDir/valley.csv};

    \nextgroupplot[
      title={Accumulated $\widehat{\Delta H}$},
      ymin=-0.08,
      ymax=0.06,
      ytick={-0.08,-0.04,0,0.04},
    ]
      \addplot[
        PeakColor,
        very thick,
        solid,
        no marks
      ]
        table[
          x=gradient_step,
          y=cumulative_first_order_entropy_change,
          col sep=comma
        ]
        {\PeakValleyDataDir/peak.csv};

      \addplot[
        ValleyColor,
        very thick,
        solid,
        no marks
      ]
        table[
          x=gradient_step,
          y=cumulative_first_order_entropy_change,
          col sep=comma
        ]
        {\PeakValleyDataDir/valley.csv};

  \end{groupplot}

  \path (group c1r1.south west) --
    node[
      anchor=north,
      yshift=-10mm
    ] {
      \pgfplotslegendfromname{peakvalleylegend}
    }
    (group c3r1.south east);

\end{tikzpicture}

\vspace{-3mm}
\caption{
Plots on Numina-Smol only on peak or valley tokens, following~\citet{yss2026gradient} and show reward, policy entropy, and accumulated $\widehat{\Delta H}$. Peak-only training accumulates positive predicted entropy change, followed by rising policy entropy and reward collapse, whereas valley-only training accumulates negative predicted entropy change and remains stable over the same training horizon.
}
\label{fig:peak-valley-entropy}
\vspace{-4mm}
\end{figure*}
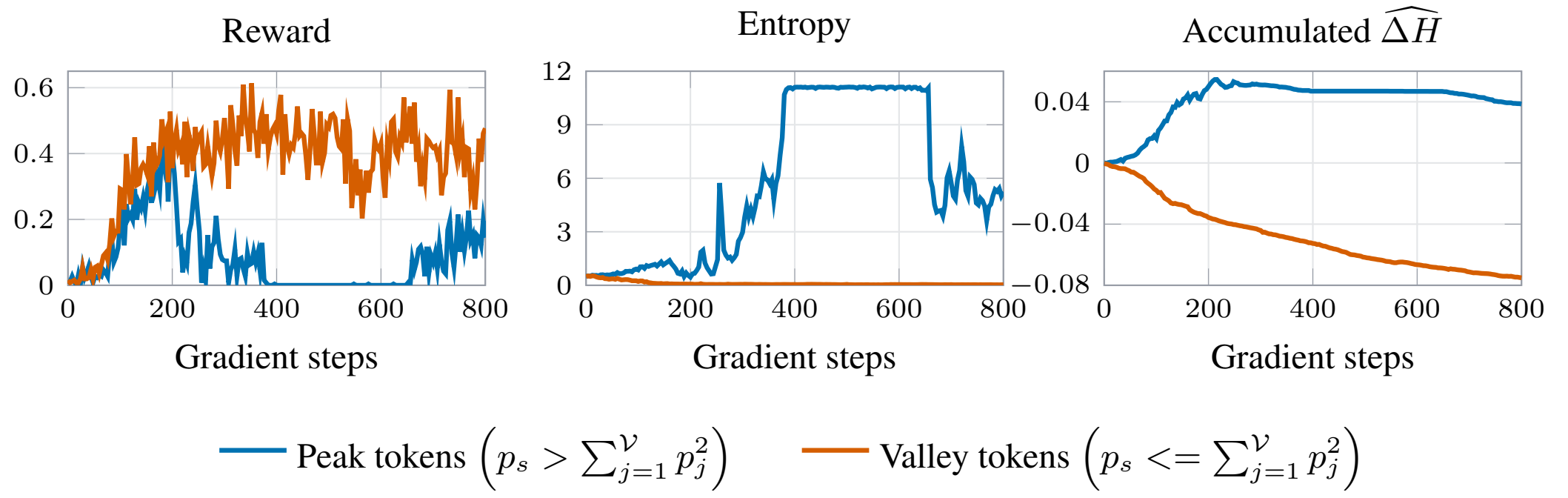
WAPO~\cite{yss2026gradient} classifies sampled tokens as \emph{peak} when $p_s>\sum_{j=1}^{\mathcal{V}} p_j^2$, where $\mathcal{V}$ is vocab size, and \emph{valley} otherwise, and reports that training only on peak tokens is unstable while training only on valley tokens remains stable. Fig.~\ref{fig:peak-valley-entropy} shows the SmolLM-3B results on NuminaMath and gives a consistent interpretation through $\widehat{\Delta H}$: peak-only training accumulates net positive predicted entropy change, followed by a sharp entropy increase and reward collapse, whereas valley-only training accumulates net negative predicted entropy change and remains stable. Peak/valley membership alone does not determine the sign of $\widehat{\Delta H}$, which also depends on the advantage. Rather, the accumulated entropy change captures the net entropy effect of retaining peak versus valley updates.


Together, these results place existing token-selection methods under a common entropy-based perspective: their effect on stability can be understood through the entropy change induced by the updates they retain. This also explains why the same token-selection strategy can provide different degrees of stabilization across training settings.

\section{Conclusion}
We study RLVR instability through the entropy change induced by policy updates. Across models and datasets, reward collapse is consistently accompanied by increasing policy entropy, and accumulated $\widehat{\Delta H}$ provides a useful first-order signal of these dynamics. Directly suppressing entropy-increasing updates improves stability, while suppressing entropy-decreasing updates accelerates collapse in collapse-prone settings. Moreover, existing token-selection methods can be understood through the same perspective: their effectiveness is closely related to how strongly they regulate entropy growth. 

Our analysis focuses on the destabilizing effect of entropy-growth; complementary work has examined entropy depletion as a cause for sub-optimal exploration. This include clipping-induced entropy effects, and alternative mechanisms for controlling entropy dynamics in RLVR~\citep{cui2025entropy,wu2026quantile,xu-etal-2026-understanding-preventing,park2025clip,hao-etal-2026-rethinking}. Developing hyperparameter-free methods that regulate both stability and exploration remains an important direction for future work.

\bibliographystyle{unsrtnat}
\bibliography{references}

\newpage
\appendix
\section{First-order entropy change}
\label{app:first-order-entropy}

We estimate the entropy change induced by an individual policy update using
the same local logit-space analysis as in ~\citet{yss2026gradient, wang2026entropy}. Consider a
fixed context with next-token distribution
$p=\operatorname{softmax}(z)$, and let $s$ denote the sampled token. For a
token whose unclipped DAPO surrogate is active, the loss is locally
\[
\ell_s = -\mathrm{sg}(\rho_s) A\cdot log(p_s),
\]
where $A$ is the group-normalized advantage and
$\rho_s=p_s/p_s^{\mathrm{old}}$ is the importance ratio. The advantage is
treated as constant during the policy update. Since
\[
\frac{\partial\,log(p_s)}{\partial z_j}
=
\mathbf{1}[j=s]-p_j,
\]
the gradient of the token loss with respect to the logits is
\[
\frac{\partial \ell_s}{\partial z_j}
=
-\rho_s A
\left(\mathbf{1}[j=s]-p_j\right).
\]
A gradient step of size $\eta$ therefore gives the local logit update
\[
\Delta z_j
=
\eta\rho_s A
\left(\mathbf{1}[j=s]-p_j\right).
\]

For $H(p)=-\sum_j p_j\log p_j$, differentiating through the softmax gives
\[
\frac{\partial H}{\partial z_j}
=
-p_j\left(\log p_j + H(p)\right).
\]
Using the first-order expansion
$\Delta H\approx\sum_j \frac{\partial H}{\partial z_j}\Delta z_j$, we obtain
\begin{align*}
\widehat{\Delta H}
&=
-\eta\rho_s A
\sum_j
p_j\left(\log p_j+H(p)\right)
\left(\mathbf{1}[j=s]-p_j\right) \\
&=
-\eta\rho_s A
\left[
p_s\log p_s
-\sum_j p_j^2\log p_j
+
H(p)
\left(
p_s-\sum_j p_j^2
\right)
\right],
\end{align*}
which gives Eq.~\ref{eq:delta_entropy}.

Equation~\ref{eq:delta_entropy} is a local first-order estimate rather than
the exact entropy change produced by a full optimizer step. In particular,
it treats the logits at each sampled context locally. In the neural network,
parameters are shared across tokens and contexts, so gradients from different
tokens interact through the parameter update. The estimate also holds the
sampled contexts fixed and therefore does not account for the change in the
distribution of contexts generated by subsequent rollout policies.
Higher-order effects and optimizer dynamics such as momentum, adaptive
preconditioning, and weight decay are likewise omitted. We therefore use
$\widehat{\Delta H}$ primarily as an estimate of the direction and relative
magnitude of entropy change.

\subsection{Accumulated entropy change.}\label{app:accumulated_entropy_change}
At each gradient step $k$, we compute Eq.~\ref{eq:delta_entropy} for all
tokens that contribute to the DAPO objective. We use the same response mask,
group-normalized advantage, and importance ratio as the training loss;
masked tokens and tokens whose clipped surrogate contributes no gradient are
excluded. Let $\mathcal{T}_k$ denote this set of contributing tokens. We
average their predicted entropy changes,
\[
\overline{\Delta H}_k
=
\frac{1}{|\mathcal{T}_k|}
\sum_{s\in\mathcal{T}_k}
\widehat{\Delta H}_{k,s},
\]
and accumulate this quantity over gradient steps:
\[
\widehat{\Delta H}_{1:K}
=
\sum_{k=1}^{K}\overline{\Delta H}_k.
\]

We compare $\widehat{\Delta H}_{1:k}$ against the measured trainer entropy
$H_k$ at the corresponding training step. Pearson and Spearman correlations
are computed over steps up to the identified collapse point. Within each
run, correlating against $H_k-H_0$ instead gives the same coefficients:
subtracting the fixed initial entropy does not affect Pearson correlation or
the rank ordering used by Spearman correlation.

\section{Experimental Setup}\label{app:exp-setup}

\paragraph{Datasets and preprocessing.}
We evaluate our method on OTT-QA, MATH-500, and NuminaMath-LEAN. \textbf{OTT-QA}~\citep{chen2020open} is a multihop QA dataset that uses wikipedia tables and passages information to answer question.
We follow the multi-turn retrieval setting of Search-R1~\citep{jinsearch}.
At each turn, the model produces a reasoning trace and either issues a search
query or returns a final answer. We generate SFT-coldstart data from 20,000 training set questions, then construct the RL training split from another independent 10,000 examples in the original training set. Each interaction is limited to ten turns. We use an F1-based answer reward augmented with a format reward,
\[
R = \frac{R_{\mathrm{F1}} + 0.3R_{\mathrm{fmt}}}{1.3}.
\]

For \textbf{MATH-500}~\citep{lightman2024let}, we use the augmented
PRM800K split with 12,000 samples as training dataset. For
\textbf{NuminaMath-LEAN}~\citep{wang2025kimina}, we filter the original
104,155-example training split to retain non-empty problems with simple numeric
answers. This procedure yields 21,251 examples, of which 20,251 are used for
training. Both datasets are formatted as single-turn
prompts. The required response format consists of a reasoning segment followed
by \texttt{Answer:}. We assign a binary reward of one when the response follows
this format and the extracted answer exactly matches the reference. All other
responses receive zero reward.

\paragraph{Importance Ratio clipping.}
In our experiments, we pick the importance ratio clipping range that make the model prone to collapse, as shown in Table \ref{tab:ir_clip_hparams}

\begin{table}[b]
  \centering
  \small
  \setlength{\tabcolsep}{5pt}
  \begin{tabular}{cllc}
    \toprule
    Exp. & Dataset & Base model & $(\epsilon_{low}, \epsilon_{high})$ \\
    \midrule
    1 & \texttt{OTT-QA} & \texttt{Qwen3-4B} & \texttt{(0.9, 9.0)} \\
    2 & \texttt{Math} & \texttt{Gemma3-4B} & \texttt{(0.2, 9.0)} \\
    3 & \texttt{Numina} & \texttt{Qwen3-4B} & \texttt{(0.9, 9.0)} \\
    4 & \texttt{Numina} & \texttt{Gemma3-4B} & \texttt{(0.2, 9.0)} \\
    \bottomrule
  \end{tabular}
  \vspace{1cm}
  \caption{Importance ratio clipping thresholds used in the four main
  experiments.}
  \label{tab:ir_clip_hparams}
\end{table}

\paragraph{Training hyperparameters.}
Unless stated otherwise, all experiments use LoRA with rank 8 and scaling
factor 32, a learning rate of $1\times10^{-5}$, off-policy steps $4$, 256 rollouts per batch, a micro-batch size of 1, a mini-batch size of 64, eight rollouts per prompt, a maximum sequence length of 8,192 tokens, and gradient clipping at norm 1.0. Math experiments use a maximum response length of 4,096 tokens. OTT-QA uses a maximum response length of 1,024 tokens, at most ten interaction turns, and a retriever with $topk=3$.

\section{Additional Experiments}\label{app:additional-exps}

\begin{figure*}[t]
\centering

\definecolor{ThresholdBlue}{HTML}{0072B2}
\definecolor{ThresholdOrange}{HTML}{E69F00}
\definecolor{ThresholdSky}{HTML}{56B4E9}
\definecolor{ThresholdRed}{HTML}{D55E00}
\definecolor{ThresholdGreen}{HTML}{009E73}
\definecolor{ThresholdGray}{HTML}{6B7280}

\providecommand{\ThresholdSeedDataDir}
  {plots/ott-qwen-threshold-seeds/data}

\pgfmathsetlengthmacro{\ThresholdSeedPanelWidth}{0.35\textwidth}
\pgfmathsetlengthmacro{\ThresholdSeedPanelHeight}{0.25\textwidth}

\pgfplotsset{
  threshold baseline/.style={
    ThresholdBlue,
    very thick,
    solid,
    mark=none
  },
  threshold 01/.style={
    ThresholdOrange,
    very thick,
    solid,
    mark=none
  },
  threshold 02/.style={
    ThresholdSky,
    very thick,
    solid,
    mark=none
  },
  threshold 02 01/.style={
    ThresholdGreen,
    very thick,
    solid,
    mark=none
  },
  threshold 01 01/.style={
    ThresholdRed,
    very thick,
    solid,
    mark=none
  },
}

\begin{subcaptiongroup}

\phantomsubcaption\label{fig:threshold-seeds:42-reward}
\phantomsubcaption\label{fig:threshold-seeds:42-entropy}
\phantomsubcaption\label{fig:threshold-seeds:42-deltah}

\phantomsubcaption\label{fig:threshold-seeds:43-reward}
\phantomsubcaption\label{fig:threshold-seeds:43-entropy}
\phantomsubcaption\label{fig:threshold-seeds:43-deltah}

\phantomsubcaption\label{fig:threshold-seeds:44-reward}
\phantomsubcaption\label{fig:threshold-seeds:44-entropy}
\phantomsubcaption\label{fig:threshold-seeds:44-deltah}

\begin{tikzpicture}
  \begin{groupplot}[
    group style={
      group size=3 by 3,
      horizontal sep=0.9cm,
      vertical sep=2.15cm,
    },
    width=\ThresholdSeedPanelWidth,
    height=\ThresholdSeedPanelHeight,
    xmin=0,
    xmax=1400,
    xtick={0,350,700,1050,1400},
    xlabel={Gradient steps},
    unbounded coords=jump,
    grid=major,
    grid style={ThresholdGray!18},
    axis line style={ThresholdGray!70},
    tick style={ThresholdGray!70},
    tick label style={font=\scriptsize},
    label style={font=\small},
    title style={
      font=\small,
      yshift=-0.5ex
    },
    scaled y ticks=false,
    yticklabel style={
      font=\scriptsize,
      /pgf/number format/fixed,
      /pgf/number format/precision=3,
      /pgf/number format/skip 0.=false,
    },
    legend style={
      draw=none,
      fill=none,
      font=\scriptsize,
      legend columns=-1,
      /tikz/every even column/.append style={
        column sep=0.65em
      },
    },
  ]


    \nextgroupplot[
      title={(\subref{fig:threshold-seeds:42-reward}) Reward},
      ymin=0,
      ymax=0.72,
      ytick={0,0.2,0.4,0.6},
      legend to name=thresholdseed42legend,
    ]
      \addplot[threshold baseline]
        table[
          x=gradient_step,
          y=reward,
          col sep=comma
        ]
        {\ThresholdSeedDataDir/seed42-baseline.csv};
      \addlegendentry{Baseline}

      \addplot[threshold 01]
        table[
          x=gradient_step,
          y=reward,
          col sep=comma
        ]
        {\ThresholdSeedDataDir/seed42-01.csv};
      \addlegendentry{$\widehat{\Delta H}<0.1$}

      \addplot[threshold 02]
        table[
          x=gradient_step,
          y=reward,
          col sep=comma
        ]
        {\ThresholdSeedDataDir/seed42-02.csv};
      \addlegendentry{$\widehat{\Delta H}<0.2$}

      \addplot[threshold 02 01]
        table[
          x=gradient_step,
          y=reward,
          col sep=comma
        ]
        {\ThresholdSeedDataDir/seed42-02-01.csv};
      \addlegendentry{$-0.2<\widehat{\Delta H}<0.1$}

      \addplot[threshold 01 01]
        table[
          x=gradient_step,
          y=reward,
          col sep=comma
        ]
        {\ThresholdSeedDataDir/seed42-01-01.csv};
      \addlegendentry{$-0.1<\widehat{\Delta H}<0.1$}


    \nextgroupplot[
      title={(\subref{fig:threshold-seeds:42-entropy}) Entropy},
      ymin=0,
      ymax=12,
      ytick={0,3,6,9,12},
    ]
      \addplot[threshold baseline]
        table[
          x=gradient_step,
          y=entropy,
          col sep=comma
        ]
        {\ThresholdSeedDataDir/seed42-baseline.csv};

      \addplot[threshold 01]
        table[
          x=gradient_step,
          y=entropy,
          col sep=comma
        ]
        {\ThresholdSeedDataDir/seed42-01.csv};

      \addplot[threshold 02]
        table[
          x=gradient_step,
          y=entropy,
          col sep=comma
        ]
        {\ThresholdSeedDataDir/seed42-02.csv};

      \addplot[threshold 02 01]
        table[
          x=gradient_step,
          y=entropy,
          col sep=comma
        ]
        {\ThresholdSeedDataDir/seed42-02-01.csv};

      \addplot[threshold 01 01]
        table[
          x=gradient_step,
          y=entropy,
          col sep=comma
        ]
        {\ThresholdSeedDataDir/seed42-01-01.csv};


    \nextgroupplot[
      title={(\subref{fig:threshold-seeds:42-deltah})
             Accumulated $\widehat{\Delta H}$},
      ymin=-550,
      ymax=100,
      ytick={-500,-300,-100,100},
      extra description/.code={
        \node[
          anchor=north west,
          font=\tiny,
          inner sep=0pt
        ] at (rel axis cs:0.02,1.15)
          {$\times 10^{-3}$};
      },
    ]
      \addplot[threshold baseline]
        table[
          x=gradient_step,
          y=cumulative_first_order_entropy_change_x1e3,
          col sep=comma
        ]
        {\ThresholdSeedDataDir/seed42-baseline.csv};

      \addplot[threshold 01]
        table[
          x=gradient_step,
          y=cumulative_first_order_entropy_change_x1e3,
          col sep=comma
        ]
        {\ThresholdSeedDataDir/seed42-01.csv};

      \addplot[threshold 02]
        table[
          x=gradient_step,
          y=cumulative_first_order_entropy_change_x1e3,
          col sep=comma
        ]
        {\ThresholdSeedDataDir/seed42-02.csv};

      \addplot[threshold 02 01]
        table[
          x=gradient_step,
          y=cumulative_first_order_entropy_change_x1e3,
          col sep=comma
        ]
        {\ThresholdSeedDataDir/seed42-02-01.csv};

      \addplot[threshold 01 01]
        table[
          x=gradient_step,
          y=cumulative_first_order_entropy_change_x1e3,
          col sep=comma
        ]
        {\ThresholdSeedDataDir/seed42-01-01.csv};


    \nextgroupplot[
      title={(\subref{fig:threshold-seeds:43-reward}) Reward},
      ymin=0,
      ymax=0.72,
      ytick={0,0.2,0.4,0.6},
      legend to name=thresholdseed43legend,
    ]
      \addplot[threshold baseline]
        table[
          x=gradient_step,
          y=reward,
          col sep=comma
        ]
        {\ThresholdSeedDataDir/seed43-baseline.csv};
      \addlegendentry{Baseline}

      \addplot[threshold 01]
        table[
          x=gradient_step,
          y=reward,
          col sep=comma
        ]
        {\ThresholdSeedDataDir/seed43-01.csv};
      \addlegendentry{$\widehat{\Delta H}<0.1$}

      \addplot[threshold 01 01]
        table[
          x=gradient_step,
          y=reward,
          col sep=comma
        ]
        {\ThresholdSeedDataDir/seed43-01-01.csv};
      \addlegendentry{$-0.1<\widehat{\Delta H}<0.1$}


    \nextgroupplot[
      title={(\subref{fig:threshold-seeds:43-entropy}) Entropy},
      ymin=0,
      ymax=12,
      ytick={0,3,6,9,12},
    ]
      \addplot[threshold baseline]
        table[
          x=gradient_step,
          y=entropy,
          col sep=comma
        ]
        {\ThresholdSeedDataDir/seed43-baseline.csv};

      \addplot[threshold 01]
        table[
          x=gradient_step,
          y=entropy,
          col sep=comma
        ]
        {\ThresholdSeedDataDir/seed43-01.csv};

      \addplot[threshold 01 01]
        table[
          x=gradient_step,
          y=entropy,
          col sep=comma
        ]
        {\ThresholdSeedDataDir/seed43-01-01.csv};


    \nextgroupplot[
      title={(\subref{fig:threshold-seeds:43-deltah})
             Accumulated $\widehat{\Delta H}$},
      ymin=-550,
      ymax=100,
      ytick={-500,-300,-100,100},
      extra description/.code={
        \node[
          anchor=north west,
          font=\tiny,
          inner sep=0pt
        ] at (rel axis cs:0.02,1.15)
          {$\times 10^{-3}$};
      },
    ]
      \addplot[threshold baseline]
        table[
          x=gradient_step,
          y=cumulative_first_order_entropy_change_x1e3,
          col sep=comma
        ]
        {\ThresholdSeedDataDir/seed43-baseline.csv};

      \addplot[threshold 01]
        table[
          x=gradient_step,
          y=cumulative_first_order_entropy_change_x1e3,
          col sep=comma
        ]
        {\ThresholdSeedDataDir/seed43-01.csv};

      \addplot[threshold 01 01]
        table[
          x=gradient_step,
          y=cumulative_first_order_entropy_change_x1e3,
          col sep=comma
        ]
        {\ThresholdSeedDataDir/seed43-01-01.csv};


    \nextgroupplot[
      title={(\subref{fig:threshold-seeds:44-reward}) Reward},
      ymin=0,
      ymax=0.72,
      ytick={0,0.2,0.4,0.6},
      legend to name=thresholdseed44legend,
    ]
      \addplot[threshold baseline]
        table[
          x=gradient_step,
          y=reward,
          col sep=comma
        ]
        {\ThresholdSeedDataDir/seed44-baseline.csv};
      \addlegendentry{Baseline}

      \addplot[threshold 01]
        table[
          x=gradient_step,
          y=reward,
          col sep=comma
        ]
        {\ThresholdSeedDataDir/seed44-01.csv};
      \addlegendentry{$\widehat{\Delta H}<0.1$}

      \addplot[threshold 01 01]
        table[
          x=gradient_step,
          y=reward,
          col sep=comma
        ]
        {\ThresholdSeedDataDir/seed44-01-01.csv};
      \addlegendentry{$-0.1<\widehat{\Delta H}<0.1$}


    \nextgroupplot[
      title={(\subref{fig:threshold-seeds:44-entropy}) Entropy},
      ymin=0,
      ymax=12,
      ytick={0,3,6,9,12},
    ]
      \addplot[threshold baseline]
        table[
          x=gradient_step,
          y=entropy,
          col sep=comma
        ]
        {\ThresholdSeedDataDir/seed44-baseline.csv};

      \addplot[threshold 01]
        table[
          x=gradient_step,
          y=entropy,
          col sep=comma
        ]
        {\ThresholdSeedDataDir/seed44-01.csv};

      \addplot[threshold 01 01]
        table[
          x=gradient_step,
          y=entropy,
          col sep=comma
        ]
        {\ThresholdSeedDataDir/seed44-01-01.csv};


    \nextgroupplot[
      title={(\subref{fig:threshold-seeds:44-deltah})
             Accumulated $\widehat{\Delta H}$},
      ymin=-550,
      ymax=100,
      ytick={-500,-300,-100,100},
      extra description/.code={
        \node[
          anchor=north west,
          font=\tiny,
          inner sep=0pt
        ] at (rel axis cs:0.02,1.15)
          {$\times 10^{-3}$};
      },
    ]
      \addplot[threshold baseline]
        table[
          x=gradient_step,
          y=cumulative_first_order_entropy_change_x1e3,
          col sep=comma
        ]
        {\ThresholdSeedDataDir/seed44-baseline.csv};

      \addplot[threshold 01]
        table[
          x=gradient_step,
          y=cumulative_first_order_entropy_change_x1e3,
          col sep=comma
        ]
        {\ThresholdSeedDataDir/seed44-01.csv};

      \addplot[threshold 01 01]
        table[
          x=gradient_step,
          y=cumulative_first_order_entropy_change_x1e3,
          col sep=comma
        ]
        {\ThresholdSeedDataDir/seed44-01-01.csv};

  \end{groupplot}


  \node[
    rotate=90,
    font=\small\bfseries,
    anchor=south
  ] at ([xshift=-1.15cm]group c1r1.west)
    {Seed 42};

  \node[
    rotate=90,
    font=\small\bfseries,
    anchor=south
  ] at ([xshift=-1.15cm]group c1r2.west)
    {Seed 43};

  \node[
    rotate=90,
    font=\small\bfseries,
    anchor=south
  ] at ([xshift=-1.15cm]group c1r3.west)
    {Seed 44};


  \path (group c1r1.south west) --
    node[
      anchor=north,
      yshift=-8mm
    ] {
      \pgfplotslegendfromname{thresholdseed42legend}
    }
    (group c3r1.south east);

  \path (group c1r2.south west) --
    node[
      anchor=north,
      yshift=-8mm
    ] {
      \pgfplotslegendfromname{thresholdseed43legend}
    }
    (group c3r2.south east);

  \path (group c1r3.south west) --
    node[
      anchor=north,
      yshift=-8mm
    ] {
      \pgfplotslegendfromname{thresholdseed44legend}
    }
    (group c3r3.south east);

\end{tikzpicture}

\end{subcaptiongroup}

\caption{%
  OTT-QA--Qwen training across three seeds with entropy-change update
  filters. Each row reports reward, entropy, and accumulated
  $\widehat{\Delta H}$ over gradient steps. Seed 42 additionally includes
  the upper filter $\widehat{\Delta H}<0.2$ and the two-sided filter
  $-0.2<\widehat{\Delta H}<0.1$. Four gradient steps are taken per logged
  training step. Curves are shown through 1400 gradient steps or their
  natural endpoint, whichever occurs first.
}
\label{fig:ott-qwen-threshold-seeds}
\end{figure*}
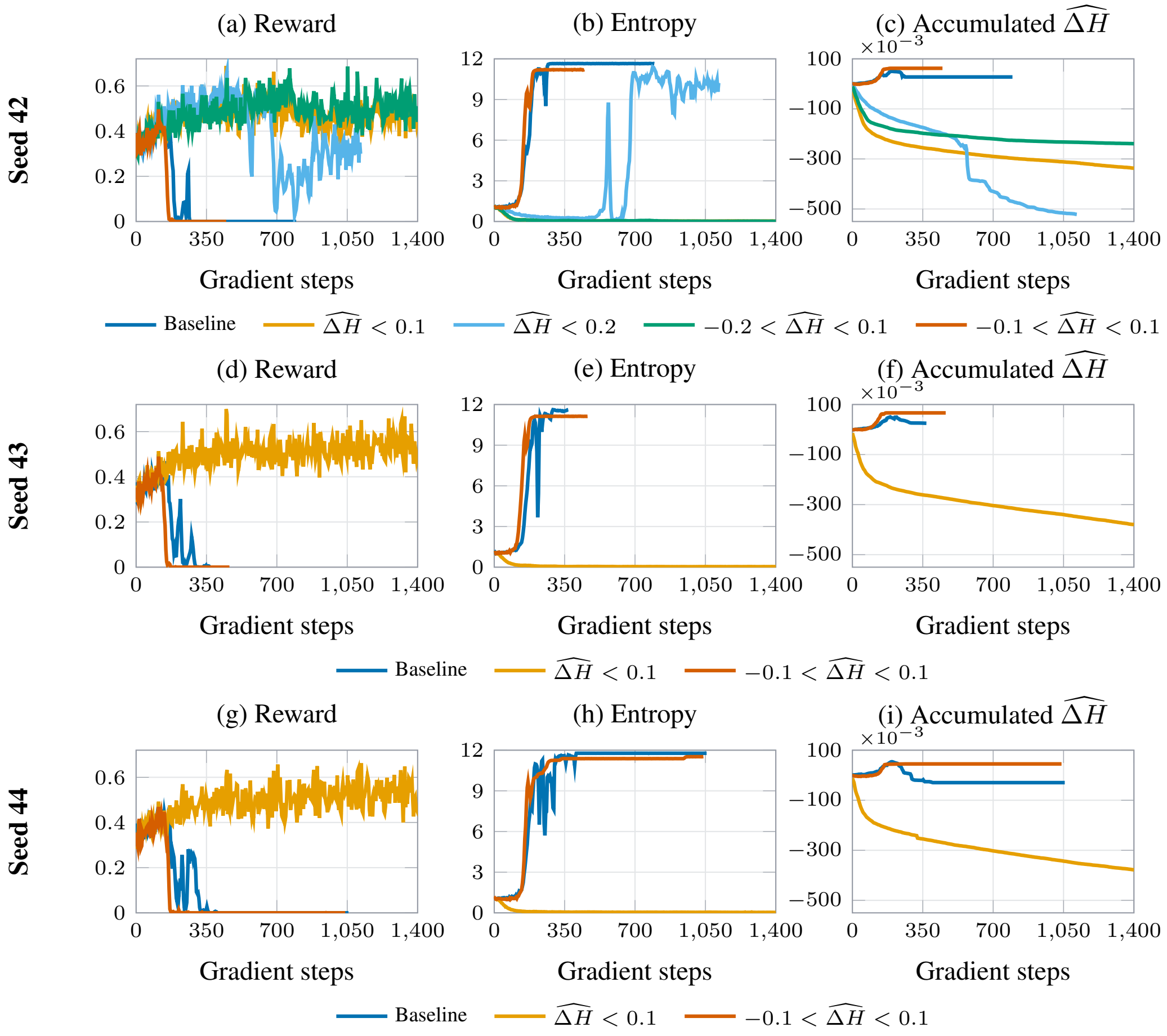

We define collapse at step $t$ if the reward falls below $50\%$ of its value
at $t$ within the following 50 gradient steps and remains below this level
for at least 10 consecutive steps. Figure~\ref{fig:threshold-seeds:42-reward}--\ref{fig:threshold-seeds:42-deltah}
shows the effect of varying the entropy-change thresholds. Increasing the
upper threshold from $\tau^+=0.1$ to $0.2$ retains more entropy-increasing
updates, resulting in a larger accumulated $\widehat{\Delta H}$ and collapse
at approximately 700 gradient steps. The lower threshold shows the opposite
intervention: with $\tau^+=0.1$ fixed, increasing $\tau^-$ from $-0.2$ to
$-0.1$ removes more entropy-decreasing updates, accelerates entropy growth,
and causes collapse at approximately 150 steps.

We repeat the experiment with two additional LoRA initialization seeds in
Fig.~\ref{fig:threshold-seeds:43-reward}--\ref{fig:threshold-seeds:43-deltah}
and Fig.~\ref{fig:threshold-seeds:44-reward}--\ref{fig:threshold-seeds:44-deltah}.
The same directional behavior is observed across seeds: allowing more
entropy-increasing updates increases accumulated $\widehat{\Delta H}$ and
reduces stability, while removing entropy-decreasing updates similarly
promotes entropy growth and earlier collapse.

%
%

%
%

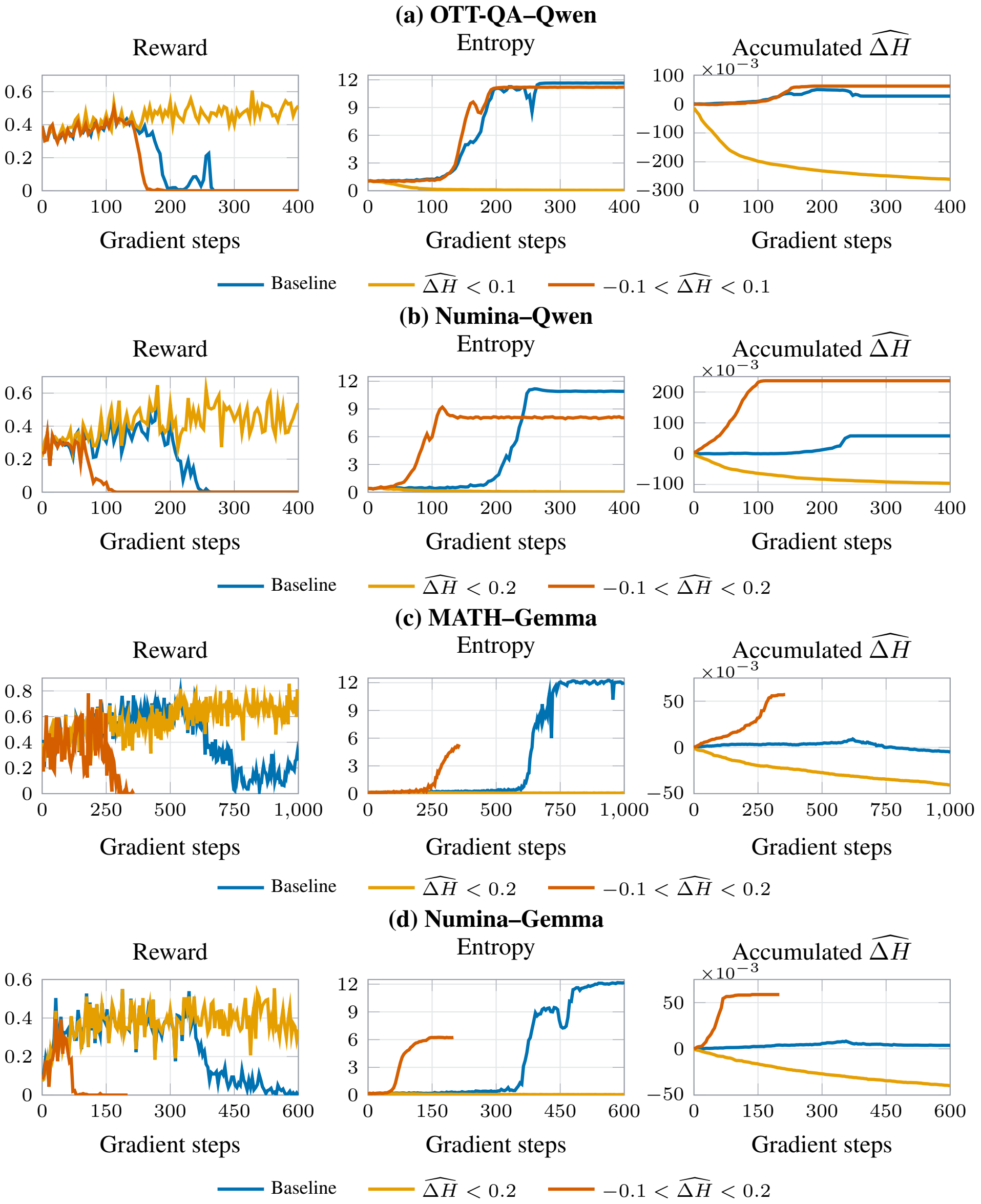
\begin{figure*}[t]
\centering

\definecolor{FilterBlue}{HTML}{0072B2}
\definecolor{FilterOrange}{HTML}{E69F00}
\definecolor{FilterRed}{HTML}{D55E00}
\definecolor{FilterGray}{HTML}{6B7280}

\providecommand{\EntropyFilteringDataDir}
  {plots/entropy-filtering-reward-entropy-full/data}

\pgfmathsetlengthmacro{\EntropyFilteringPanelWidth}{0.35\textwidth}
\pgfmathsetlengthmacro{\EntropyFilteringPanelHeight}{0.22\textwidth}

\pgfplotsset{
  entropy filter baseline/.style={
    FilterBlue,
    very thick,
    solid,
    mark=none
  },
  entropy filter upper/.style={
    FilterOrange,
    very thick,
    solid,
    mark=none
  },
  entropy filter two sided/.style={
    FilterRed,
    very thick,
    solid,
    mark=none
  },
}

\newcommand{\AddEntropyFilterCurves}[2]{%
  \addplot[entropy filter baseline]
    table[
      x=gradient_step,
      y=#2,
      col sep=comma
    ]
    {\EntropyFilteringDataDir/#1-baseline.csv};

  \addplot[entropy filter upper]
    table[
      x=gradient_step,
      y=#2,
      col sep=comma
    ]
    {\EntropyFilteringDataDir/#1-upper.csv};

  \addplot[entropy filter two sided]
    table[
      x=gradient_step,
      y=#2,
      col sep=comma
    ]
    {\EntropyFilteringDataDir/#1-two-sided.csv};
}

\begin{subcaptiongroup}

\phantomsubcaption
\label{fig:entropy-filtering-seed42:ott-qwen}

\phantomsubcaption
\label{fig:entropy-filtering-seed42:numina-qwen}

\phantomsubcaption
\label{fig:entropy-filtering-seed42:math-gemma}

\phantomsubcaption
\label{fig:entropy-filtering-seed42:numina-gemma}

\begin{tikzpicture}
  \begin{groupplot}[
    group style={
      group size=3 by 4,
      horizontal sep=0.9cm,
      vertical sep=2.4cm,
    },
    width=\EntropyFilteringPanelWidth,
    height=\EntropyFilteringPanelHeight,
    xmin=0,
    xlabel={Gradient steps},
    unbounded coords=jump,
    grid=major,
    grid style={FilterGray!18},
    axis line style={FilterGray!70},
    tick style={FilterGray!70},
    tick label style={font=\scriptsize},
    label style={font=\small},
    title style={
      font=\small,
      yshift=-0.5ex
    },
    scaled y ticks=false,
    yticklabel style={
      font=\scriptsize,
      /pgf/number format/fixed,
      /pgf/number format/precision=2,
      /pgf/number format/skip 0.=false,
    },
    legend style={
      draw=none,
      fill=none,
      font=\scriptsize,
      legend columns=-1,
      /tikz/every even column/.append style={
        column sep=0.9em
      },
    },
  ]


    \nextgroupplot[
      title={Reward},
      xmax=400,
      xtick={0,100,200,300,400},
      ymin=0,
      ymax=0.7,
      ytick={0,0.2,0.4,0.6},
      legend to name=ottqwenfilterlegend,
    ]
      \AddEntropyFilterCurves{ott-qwen}{reward}
      \addlegendentry{Baseline}
      \addlegendentry{$\widehat{\Delta H}<0.1$}
      \addlegendentry{$-0.1<\widehat{\Delta H}<0.1$}

    \nextgroupplot[
      title={Entropy},
      xmax=400,
      xtick={0,100,200,300,400},
      ymin=0,
      ymax=12.5,
      ytick={0,3,6,9,12},
    ]
      \AddEntropyFilterCurves{ott-qwen}{entropy}

    \nextgroupplot[
      title={Accumulated $\widehat{\Delta H}$},
      xmax=400,
      xtick={0,100,200,300,400},
      ymin=-300,
      ymax=100,
      ytick={-300,-200,-100,0,100},
      extra description/.code={
        \node[
          anchor=north west,
          font=\tiny,
          inner sep=0pt
        ] at (rel axis cs:0.02,1.15)
          {$\times 10^{-3}$};
      },
    ]
      \AddEntropyFilterCurves
        {ott-qwen}
        {cumulative_first_order_entropy_change_x1e3}


    \nextgroupplot[
      title={Reward},
      xmax=400,
      xtick={0,100,200,300,400},
      ymin=0,
      ymax=0.7,
      ytick={0,0.2,0.4,0.6},
      legend to name=numinaqwenfilterlegend,
    ]
      \AddEntropyFilterCurves{numina-qwen}{reward}
      \addlegendentry{Baseline}
      \addlegendentry{$\widehat{\Delta H}<0.2$}
      \addlegendentry{$-0.1<\widehat{\Delta H}<0.2$}

    \nextgroupplot[
      title={Entropy},
      xmax=400,
      xtick={0,100,200,300,400},
      ymin=0,
      ymax=12.5,
      ytick={0,3,6,9,12},
    ]
      \AddEntropyFilterCurves{numina-qwen}{entropy}

    \nextgroupplot[
      title={Accumulated $\widehat{\Delta H}$},
      xmax=400,
      xtick={0,100,200,300,400},
      ymin=-125,
      ymax=250,
      ytick={-100,0,100,200},
      extra description/.code={
        \node[
          anchor=north west,
          font=\tiny,
          inner sep=0pt
        ] at (rel axis cs:0.02,1.15)
          {$\times 10^{-3}$};
      },
    ]
      \AddEntropyFilterCurves
        {numina-qwen}
        {cumulative_first_order_entropy_change_x1e3}


    \nextgroupplot[
      title={Reward},
      xmax=1000,
      xtick={0,250,500,750,1000},
      ymin=0,
      ymax=0.9,
      ytick={0,0.2,0.4,0.6,0.8},
      legend to name=mathgemmafilterlegend,
    ]
      \AddEntropyFilterCurves{math-gemma}{reward}
      \addlegendentry{Baseline}
      \addlegendentry{$\widehat{\Delta H}<0.2$}
      \addlegendentry{$-0.1<\widehat{\Delta H}<0.2$}

    \nextgroupplot[
      title={Entropy},
      xmax=1000,
      xtick={0,250,500,750,1000},
      ymin=0,
      ymax=12.5,
      ytick={0,3,6,9,12},
    ]
      \AddEntropyFilterCurves{math-gemma}{entropy}

    \nextgroupplot[
      title={Accumulated $\widehat{\Delta H}$},
      xmax=1000,
      xtick={0,250,500,750,1000},
      ymin=-50,
      ymax=75,
      ytick={-50,0,50},
      extra description/.code={
        \node[
          anchor=north west,
          font=\tiny,
          inner sep=0pt
        ] at (rel axis cs:0.02,1.15)
          {$\times 10^{-3}$};
      },
    ]
      \AddEntropyFilterCurves
        {math-gemma}
        {cumulative_first_order_entropy_change_x1e3}


    \nextgroupplot[
      title={Reward},
      xmax=600,
      xtick={0,150,300,450,600},
      ymin=0,
      ymax=0.6,
      ytick={0,0.2,0.4,0.6},
      legend to name=numinagemmafilterlegend,
    ]
      \AddEntropyFilterCurves{numina-gemma}{reward}
      \addlegendentry{Baseline}
      \addlegendentry{$\widehat{\Delta H}<0.2$}
      \addlegendentry{$-0.1<\widehat{\Delta H}<0.2$}

    \nextgroupplot[
      title={Entropy},
      xmax=600,
      xtick={0,150,300,450,600},
      ymin=0,
      ymax=12.5,
      ytick={0,3,6,9,12},
    ]
      \AddEntropyFilterCurves{numina-gemma}{entropy}

    \nextgroupplot[
      title={Accumulated $\widehat{\Delta H}$},
      xmax=600,
      xtick={0,150,300,450,600},
      ymin=-50,
      ymax=75,
      ytick={-50,0,50},
      extra description/.code={
        \node[
          anchor=north west,
          font=\tiny,
          inner sep=0pt
        ] at (rel axis cs:0.02,1.15)
          {$\times 10^{-3}$};
      },
    ]
      \AddEntropyFilterCurves
        {numina-gemma}
        {cumulative_first_order_entropy_change_x1e3}

  \end{groupplot}


  \path (group c1r1.north west) --
    node[
      anchor=south,
      yshift=5mm,
      font=\small\bfseries
    ] {
      (\subref{fig:entropy-filtering-seed42:ott-qwen})
      OTT-QA--Qwen
    }
    (group c3r1.north east);

  \path (group c1r2.north west) --
    node[
      anchor=south,
      yshift=5mm,
      font=\small\bfseries
    ] {
      (\subref{fig:entropy-filtering-seed42:numina-qwen})
      Numina--Qwen
    }
    (group c3r2.north east);

  \path (group c1r3.north west) --
    node[
      anchor=south,
      yshift=5mm,
      font=\small\bfseries
    ] {
      (\subref{fig:entropy-filtering-seed42:math-gemma})
      MATH--Gemma
    }
    (group c3r3.north east);

  \path (group c1r4.north west) --
    node[
      anchor=south,
      yshift=5mm,
      font=\small\bfseries
    ] {
      (\subref{fig:entropy-filtering-seed42:numina-gemma})
      Numina--Gemma
    }
    (group c3r4.north east);


  \path (group c1r1.south west) --
    node[
      anchor=north,
      yshift=-8mm
    ] {
      \pgfplotslegendfromname{ottqwenfilterlegend}
    }
    (group c3r1.south east);

  \path (group c1r2.south west) --
    node[
      anchor=north,
      yshift=-8mm
    ] {
      \pgfplotslegendfromname{numinaqwenfilterlegend}
    }
    (group c3r2.south east);

  \path (group c1r3.south west) --
    node[
      anchor=north,
      yshift=-8mm
    ] {
      \pgfplotslegendfromname{mathgemmafilterlegend}
    }
    (group c3r3.south east);

  \path (group c1r4.south west) --
    node[
      anchor=north,
      yshift=-8mm
    ] {
      \pgfplotslegendfromname{numinagemmafilterlegend}
    }
    (group c3r4.south east);

\end{tikzpicture}

\end{subcaptiongroup}

\caption{%
  Seed-42 training results for the baseline, upper entropy-change
  filter, and two-sided entropy-change filter. The precise filtering
  intervals are reported in each row's legend. Each row corresponds
  to one dataset--model setting and reports reward, policy entropy,
  and accumulated $\widehat{\Delta H}$. The horizontal axes report
  gradient steps, with four gradient steps per logged training step.
  Curves that terminate before the displayed limit stop at their
  final logged point.
}
\label{fig:entropy-filtering-seed42}
\end{figure*}



\end{document}